\documentclass[10pt,british,10pt,twocolumn,letterpaper]{article}
\usepackage[T1]{fontenc}
\usepackage[latin9]{inputenc}
\usepackage{verbatim}
\usepackage{units}
\usepackage{amstext}
\usepackage{graphicx}

\makeatletter

\providecommand{\tabularnewline}{\\}

\usepackage{cvpr}
\usepackage{times}
\usepackage{epsfig}
\usepackage{graphicx}
\usepackage{amsmath}
\usepackage{amssymb}

\usepackage{placeins}
\usepackage{hyphenat}
\usepackage{xspace}

\usepackage{color}
\usepackage{colortbl}
\usepackage{rotating}
\definecolor{lightgray}{gray}{0.8}
\definecolor{verylightgray}{gray}{0.9}

\usepackage[pagebackref=true,breaklinks=true,letterpaper=true,colorlinks,bookmarks=false]{hyperref}

\cvprfinalcopy 

\def\cvprPaperID{{6117}\vspace{-1em}} 

\ifcvprfinal\fi

\makeatother

\usepackage{babel}
\begin{document}
\makeatletter
\renewcommand{\paragraph}{%
\@startsection{paragraph}{4}%
 {\z@}{0.5ex \@plus 1ex \@minus .2ex}{-0.5em}%
  {\normalfont \normalsize \bfseries}%
}
\makeatother

\setlength{\textfloatsep}{1.5em}

\let\originalparagraph\paragraph
\renewcommand{\paragraph}[2][.]{\vspace{0.0em}\originalparagraph{#2#1}}

\title{\vspace{-2em}Large-scale interactive object segmentation with human annotators\vspace{-1.0em}}
\author{Rodrigo Benenson\hspace{-7em}
\and Stefan Popov\vspace{0.25em}\\
\begin{tabular}{c}
Google Research\tabularnewline
\texttt{\small{}\{benenson, spopov, vittoferrari\}@google.com}
\vspace{-0.5em}
\end{tabular}
\and\hspace{-7em}Vittorio Ferrari}

\maketitle
\vspace{-1em}
\begin{abstract}
\vspace{-0.5em}
Manually annotating object segmentation masks is very time consuming.
Interactive object segmentation methods offer a more efficient alternative where
a human annotator and a machine segmentation model collaborate.
In this paper we make several contributions to interactive segmentation:
(1) we systematically explore in simulation the design space of deep interactive segmentation
models and report new insights and caveats;
(2) we execute a large-scale annotation campaign with real human annotators, producing masks for 2.5M instances on the OpenImages dataset.
We plan to release this data publicly, forming the largest existing dataset for instance segmentation. Moreover, by re-annotating part of the COCO dataset, we show that we can produce instance masks $3\times$ faster than traditional polygon drawing tools while also providing better quality.
(3) We present a technique for automatically estimating the quality of the produced masks which exploits indirect signals from the annotation process.
\end{abstract}

\vspace{-1.5em}
\section{\label{sec:Introduction}Introduction}
\vspace{-0.5em}

Propelled by the increased computing power, the last years have seen
a dramatic growth in the size of models for computer vision tasks. These larger
models are evermore demanding of larger training sets to reach performance
saturation \cite{Sun2017Iccv}. This demand for data often becomes a
bottleneck for practitioners. While computers become cheaper, faster,
and able to handle larger models, the cost of humans manually
annotating data remains very high. Hence, we need new
strategies to scale-up human annotations.

Amongst the traditional image understanding tasks, instance segmentation is considered one
of the most expensive to annotate~\cite{Everingham2015Ijcv,Lin2014EccvCoco,Cordts2016Cvpr,Zhou2017Cvpr}.
For each object instance in each class of interest, it requires annotating a mask indicating
which pixels belong to the instance.

In this work we explore an interactive segmentation approach to annotate instance
masks, where the human annotator focuses on correcting the output of
a segmentation model, rather than spending time blindly creating full
annotations that might be redundant or already captured by the model.
Across multiple rounds the annotator provides corrections to the current segmentation,
and then the model incorporates them to refine the segmentation.

\begin{figure}
\hfill{}\includegraphics[width=1\columnwidth,height=0.34\columnwidth]{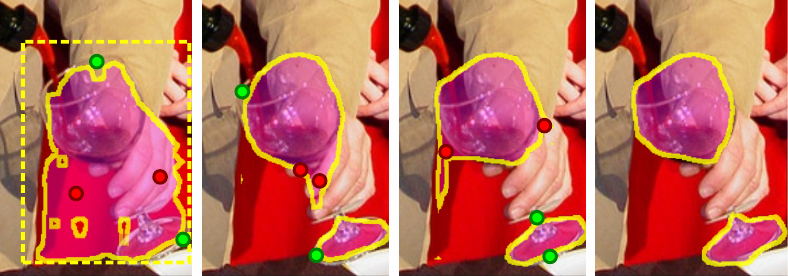}\hfill{}
\vspace{-1em}
\caption{\label{fig:amazing-corrective-clicks}Example of corrective clicks
and their effect on the segmentation mask. Starting from a bounding box,
the annotator provides up to $4$ corrective clicks in each round. See section \ref{sec:framework}.}
\vspace{-0.5em}
\end{figure}

Albeit the idea of interactive segmentation was established already
a decade ago \cite{Boykov2001Iccv,Rother2004AcmtgGrabCut},
we make two key contributions:
(1) through extensive simulations we systematically explore the design space of deep interactive segmentations
models and report new insights and caveats (Sec.~\ref{sec:cc-simulations}).
(2) while most previous works report only simulation experiments \cite{Xu2016Eccv,Xu2017Bmvc,Liew2017Iccv,Mahadevan2018Bmvc,Hu2018Nn,Maninis2018Cvpr},
we execute a large-scale annotation campaign with real human annotators (2.5M instances, Sec.~\ref{sec:annotation-process}),
we analyse the annotators behavior (Sec.~\ref{sec:do-annotators-behave}) and the resulting annotations
(Sec.~\ref{sec:time-vs-masks-quality}).

Our results show that
we can make instance segmentation $3\times$ faster than traditional polygon
drawing tools~\cite{Lin2014EccvCoco} while also providing better quality (Sec.~\ref{sec:time-vs-masks-quality}).
Additionally our method can produce masks across different time budgets, and
we present a new technique for automatically estimating the quality of the produced masks
which exploits indirect signals from the annotation process (Sec.~\ref{sec:is-the-ranking-good}).

To the best of our knowledge this is the first work exploring interactive
annotations at scale. We apply our approach to collect $2.5\text{M}$
new masks over $300$ categories of the OpenImages dataset \cite{Kuznetsova2018OpenImages}.
We plan to release the data mid-2019, making it the largest public dataset for instance segmentation (in the number of instances).

\begin{figure*}
\vspace{-1.5em}
\hfill{}\includegraphics[width=0.75\paperwidth,height=0.55\columnwidth]{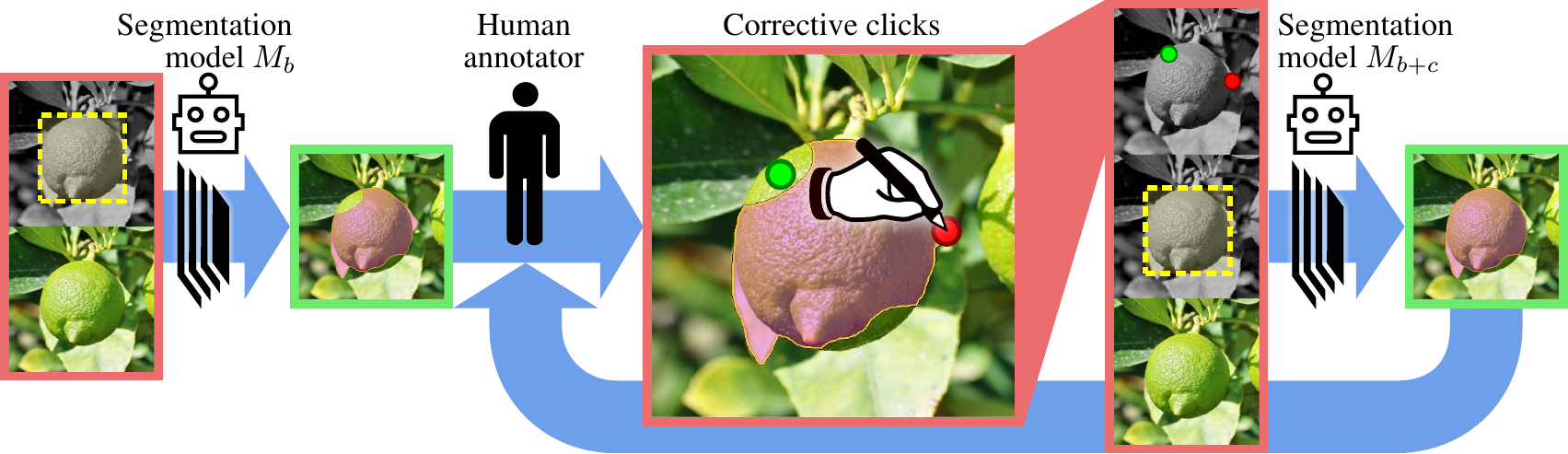}\hfill{}
\vspace{-0.2em}
\caption{\label{fig:system}High level overview of our interactive annotation
system ($M_{r}$ not shown). See description in section \ref{sec:framework}.}
\vspace{-1.25em}
\end{figure*}

\vspace{-0.25em}
\subsection{\label{subsec:Related-work}Related work}
\vspace{-0.25em}

\paragraph{Dataset annotations}

A flurry of techniques have been explored to annotate the location
of objects in images: bounding box annotations \cite{Everingham2015Ijcv},
clicks on the object extremes \cite{Papadopoulos2017Iccv,Maninis2018Cvpr},
clicks on the object centre \cite{Bearman2016Eccv,Papadopoulos2017Eccv},
bounding box followed by edits to a generated segment \cite{Boykov2001Iccv,Rother2004AcmtgGrabCut},
scribbles \cite{Bearman2016Eccv,Lin2016Cvpr}, %
hierarchical super-pixel annotations \cite{Maire2013Bmvc}, polygon
annotations \cite{Everingham2015Ijcv,Bharath2011Iccv,Lin2014EccvCoco},
interactive polygons \cite{Acuna2018Cvpr}, eye gaze \cite{Papadopoulos2014Eccv},
and touch interfaces \cite{Pizenberg2017ACMMC}, just to name a few.\\
Despite many explored ideas, the current most popular datasets for
object instances localization were all obtained using purely manual
annotations of boxes or (layered) polygons \cite{Everingham2015Ijcv,Russakovsky2015IjcvImageNet,Bharath2011Iccv,Lin2014EccvCoco,Cordts2016Cvpr,Neuhold2017Iccv,Huang2018ArxivApolloscape,Zhou2017Cvpr}.%
\\
This work aims at changing the status quo by showing the interest
of interactive instance segmentation as main annotation
strategy for large scale instance segmentation.

\paragraph{Weakly supervised segmentation}

Weakly supervised segmentation methods reconstruct approximate segmentation
masks starting from weaker supervision input. For semantic/instance
segmentation the following sources of supervision have been explored:
image level label only \cite{Zhou2018Cvpr}, point clicks \cite{Bearman2016Eccv,Xu2016Eccv},
boxes only \cite{Dai2015Iccv,Khoreva2017Cvpr,Xu2017Bmvc}, scribbles
only \cite{Lin2016Cvpr,Bearman2016Eccv}, boxes plus clicks \cite{Boykov2001Iccv,Rother2004AcmtgGrabCut,Xu2017Bmvc,Papadopoulos2017Iccv,Maninis2018Cvpr}%
.%
\\
Most interactive segmentation techniques build upon weakly supervised
methods. Our work starts from segments generated from boxes, and then
moves into an interactive mode by adding corrective clicks as additional supervision.

\paragraph{Interactive segmentation}

The idea of interactive image segmentation is decades old \cite{Boykov2001Iccv,Rother2004AcmtgGrabCut}.
It has been revisited in the recent years under the deep learning
umbrella \cite{Xu2016Eccv,Mahadevan2018Bmvc,Hu2018Nn,Wang2018PamiDeepigeos,Maninis2018Cvpr,Li2018Cvpr,Le2018Eccv,Acuna2018Cvpr},
focusing on how to best use clicks and scribbles corrections. Most
of these work share structural similarities but diverge in their finer
design choices (e.g. annotation type, input encoding, etc.). Section \ref{sec:cc-simulations}
revisits some of these choices in an unified experimental setup and
draws new conclusions.

Despite the theme of human-machine interaction, most previous work
in this area report purely simulation experiments
\cite{Xu2017Bmvc,Liew2017Iccv,Mahadevan2018Bmvc,Maninis2018Cvpr}
or only small scale human experiments ($\leq50$ instances \cite{Le2018Eccv,Li2018Cvpr,Acuna2018Cvpr}).
We instead collect $2.5\text{M}$ masks over 300 categories
using a team of 100 annotators, share the learned lessons,
and analyse the resulting annotations when considering interactive
segmentation at scale.

\section{\label{sec:framework}Overall system design}

We propose a design that involves three types of models
$M_{b}$, $M_{b+c}$, $M_{r}$.
Starting from existing bounding boxes for each instance, we generate
initial masks via an \textquotedbl{}image$+$box$\rightarrow$mask\textquotedbl{}
model $M_{b}$. We then
show these masks to human annotators, which
indicate \emph{corrections}. These corrections
are used as supervision to generate new improved masks, via an \textquotedbl{}image$+$box$+$corrections$\rightarrow$mask\textquotedbl{}
model $M_{b+c}$. The annotators iteratively correct the outputs of
$M_{b+c}$ over multiple rounds (figure \ref{fig:system}).
The final instance segmentation model is thus trained using as privileged
information the initial bounding boxes and the corrections provided throughout the process.

Models $M_{b}$ and $M_{b+c}$ are implemented using convnets, while
$M_{r}$ is a decision forest. Compared to previous work on interactive segmentation such as \cite{Mahadevan2018Bmvc,Li2018Cvpr,Acuna2018Cvpr}
we use a different design for $M_{b}$ / $M_{b+c}$, and introduce the use of a ranking model $M_{r}$.
This ranking model uses the time sequence of annotator corrections
on an instance to predict the expected quality of
the generated mask. This ranking can be used to prioritise further
corrections on low quality instances, or as weighting signal when
training an instance segmentation model.
In section \ref{sec:cc-simulations} we study the
design of $M_{b}$ and $M_{b+c}$; and in section \ref{sec:annotation-process}
we describe a concrete instantiation of our approach. The ranking
model $M_{r}$ is described in section \ref{subsec:Ranking-model}.

Since annotators spend their time solely on corrections, the data
collected is directly focused on areas not yet correctly captured
by the $M_{b}$/$M_{b+c}$ models. Easy instances will need zero edits
from the get go, while hard instances might need multiple revisits.

\section{\label{sec:cc-simulations}Simulations}

The generic system described in section \ref{sec:framework} has many
free design parameters, and previous related works have reported contradicting
results.
Section \ref{subsec:Experiments-blueprint} describes an
implementation blueprint of our system, and section \ref{sec:simulation-results}
reports simulation results exploring its design space.

\subsection{\label{subsec:Experiments-blueprint}Experiments blueprint}
\vspace{-0.25em}

\paragraph{Evaluation set}

For these experiments we use COCO \cite{Lin2014EccvCoco}, which is considered the gold
standard amongst existing instance segmentation datasets. COCO was
annotated by manually drawing polygons on object instances. Albeit some previous
interactive segmentation works report COCO results \cite{Xu2016Eccv,Liew2017Iccv,Li2018Cvpr,Hu2018Nn},
they only consider $\le20$ instances per class; here we consider
a larger set of $\sim\negmedspace88k$  instances.\\
COCO objects have a median size of $53\times62$ pixels.
Small objects tend to have blob-like shapes, while larger objects
show more detail. We thus focus training and evaluation only on instances
larger than $80\times80\ \text{pixels}$. We evaluate results over
the large instances in the COCO 2014 validation set (41k images), which
we denote as $\text{COCO}_{\text{L}}^{\text{test}}$.

\paragraph{Training set}

The training set of $M_{b}$ and $M_{b+c}$ will have a direct impact
on the effectiveness of the interactive annotation system.
We train our models over a subset of the ADE20k training set \cite{Zhou2017Cvpr}.
We picked the 255 largest classes representing objects (window, leg, lamp,
etc.) and left out stuff classes (wall, sky, sidewalk, etc.).
In total these have $\sim\negmedspace400k$ instances. After filtering by
size, we are left with $\sim\negmedspace134k$ instances with ground-truth segmentations.
We name this training set $\text{ADE}_{\text{L}}^{\text{train}}$.
Unless otherwise specified, we train our models on it in a class-agnostic way:
considering all instances as belonging to a single generic object class.
We split $\text{ADE}_{\text{L}}^{\text{train}}$ in two halves to
train $M_{b}$ over $\text{ADE}_{\text{L\ \ensuremath{\nicefrac{1}{2}}}}^{\text{train}}$
, and $M_{b+c}$ over $\text{ADE}_{\text{L\ \ensuremath{\nicefrac{2}{2}}}}^{\text{train}}$.\\
Unless otherwise specified all instances are cropped and scaled to
fit inside a $193\times193\ \text{pixels}$ box (keeping their aspect
ratio) centred in a $385\times385\ \text{pixels}$ image (capturing some
of the instance context, or padding borders with black pixels if outside
of source image). \\
The ranking model $M_{r}$ is trained over a small set of real corrective click
annotations and ground-truth masks (\S \ref{subsec:Ranking-model}).

We treat both training and testing on a per-instance basis%
, and ignore class labels (i.e. we average across instances). We
use the traditional mean intersection-over-union
(mIoU) as our main evaluation metric.

\paragraph{Annotator simulation}

Training $M_{b}$ requires ground-instance segmentations, and training
$M_{b+c}$ additionally requires annotator corrections.
These training corrections can be collected by running annotation campaigns over data with already available ground-truth segmentations, or can be generated by simulating the expected annotator behavior.
In the latter case, the details of the simulations matter.
The assumptions underlying the simulated corrections should match the
expected behavior of annotators. Moreover, depending on how much
error we expect from the annotators (noise model), the simulation results will have higher or lower IoU.
This is a confounding factor when comparing results amongst related works.
Also, if we train with a certain annotator noise model, and test with the same model, results might
be over-optimistic (compared to using actual human annotators).

Our simulations include noise both on the boxes and the clicks.
We generate object bounding boxes by perturbing the corners of a perfect bounding box fit to the ground-truth segmentation with $\mathcal{N}\left(0,\,60\ \text{pixels}\right)$ noise. We keep bounding boxes with $\text{IoU}\ge0.7$ with the tight ground-truth box.
We use such loose boxes because:
a) depending on their source, not all boxes are expected to be very tight;
b) we expect to encounter cases where there is a drift between the box annotation policies and
the segmentation policies (section \ref{sec:policies}).
Furthermore, we perturb corrective clicks with $\mathcal{N}\left(0,\,3\ \text{pixels}\right)$.
The initial click location is also randomly sampled following a probability distributions
specific to the click type (section \ref{sec:boundary-vs-region-click}).
Also, some error regions might be ignored if deemed too small (section \ref{sec:annotation-noise}).\\
Unless otherwise stated, each simulation runs for 3 rounds of interaction, with the simulated annotator providing up to 3 clicks per round (which we denote as $3\times3$). Section \ref{sec:number-of-clicks-and-rounds}
studies how to distribute corrective clicks across rounds.\\

\paragraph{Models}

Both $M_{b}$ and $M_{b+c}$ use the same architecture. We train Deeplabv2
ResNet101 \cite{Chen2018PamiDeeplab} for per-pixel binary classification
(instance foreground/background). See supplementary material for training
parameters.
The Deeplab model is augmented to accept N-channels input
instead of only an RGB image. $M_{b}$ uses 4 channels, RGB plus a
binary bounding box image (inside/outside). $M_{b+c}$ uses
5 or more channels, RGB + box + corrections. Previous works have used
different strategies to encode the corrections, which we explore in section \ref{sec:clicks-encoding}.
Our ranking model $M_{r}$ is a decision
forest described in section \ref{subsec:Ranking-model}.\\
By default, we train $M_{b+c}$ to handle $3\times3$ rounds.
We do this by training a single model to use as inputs the cropped RGB image,
the binary bounding box, and a variable number of clicks between
1 and 9 (random, uniform discrete distribution). Clicks are encoded as small binary disks (section \ref{sec:clicks-encoding}).
We first train the $M_{b}$ model over $\text{ADE}_{\text{L\ \ensuremath{\nicefrac{1}{2}}}}^{\text{train}}$.
Then we train $M_{b+c}$ using simulated corrective clicks on $\text{ADE}_{\text{L\ \ensuremath{\nicefrac{2}{2}}}}^{\text{train}}$ over masks generated by $M_{b}$.

\subsection{\label{sec:simulation-results}Simulation results}
\vspace{-0.25em}

Most experiments in this section require re-training both $M_{b}$
and $M_{b+c}$. Together they represent over 9 GPU-months of training time.

\subsubsection{\label{sec:M_b-baselines}$M_{b}$ baselines}
\vspace{-0.25em}

When training $M_{b}$ over $\text{ADE}_{\text{L\ \ensuremath{\nicefrac{1}{2}}}}^{\text{train}}$
we obtain a mean IoU of $65\%$ on $\text{COCO}_{\text{L}}^{\text{test}}$.
This is the starting point for our annotations. For comparison using
the raw (noisy) bounding boxes as masks, we obtain $50\%\ \text{mIoU}$.
A well-tuned Grabcut implementation reaches $59\%\ \text{mIoU}$ \cite{Papadopoulos2017Iccv}.
Overall our class-agnostic transfer from ADE20k to COCO via $M_{b}$ seems to perform well in comparison.

\subsubsection{\label{sec:boundary-vs-region-click}Boundary click or region click?}
\vspace{-0.25em}

When considering which corrective clicks should be done over a mask,
the existing literature is split between clicks at the object border
\cite{Papadopoulos2017Iccv,Maninis2018Cvpr,Le2018Eccv,Acuna2018Cvpr}
or clicks inside the error regions \cite{Xu2016Eccv,Liew2017Iccv,Li2018Cvpr}.

For boundary clicks we train $M_{b+c}$ with all clicks pasted into a single
channel, whereas region clicks are encoded in two separate channels depending if the click
was done inside or outside the current mask.

At test time, we iteratively simulate the clicks by first adding 3
corrective clicks over the mask created by $M_{b}$, then applying $M_{b+c}$,
then adding 3 additional corrective clicks, then re-applying $M_{b+c}$, etc.
The corrective clicks are applied simulating either clicks on the
boundary of the error regions or in their centre. The likelihood
of a click in a error region is proportional to its area. If multiple clicks
fall in the same error region, they are spread out to roughly partition
it in equal areas (or the boundary in equal lengths).

\paragraph{Result}

Both type of clicks bring clear improvements in every round.
After three rounds, region clicks reach $80\%\ \text{mIoU}$ while
boundary clicks reach only $77\%\ \text{mIoU}$. This trend is consistent
across different type of input encoding and number of clicks/rounds.
We thus move forward with corrective clicks in the centre of error regions as our main correction strategy.

This result can be explained via:
1) centre region clicks are more robust: after a small perturbation the click is
still roughly at the region centre, while noisy boundary clicks are
often off,
2) region clicks provide more explicit information about what needs to be added or removed.

\vspace{-0.25em}
\subsubsection{\label{sec:annotation-noise}Annotation noise}

We report the effect of two
aspects of the annotator behavior model:
(1) how precise is the click
of the annotator (click noise)?,
(2) which error regions will be considered too small to be clicked on (minimum
region size)?
If the goal is to get masks annotated as fast as possible, small error
regions should be ignored by the annotator since they will
have minor impact on mask quality.

\paragraph{Result}

We consider the mIoU reached at
the end of $3\times3$ simulations. Compared to zero click noise,
 adding an isotropic Gaussian click noise with standard deviation $3$ or $6$ pixels causes
a $3\%$ and $7\%$ drop in mIoU, respectively. Similarly,
if the annotator ignores regions smaller than
$x^{2}$ pixels, we observe a drop of
$3\%$ or $8\%$ mIoU at $x=20$ or $x=30$, respectively (using click noise 3 pixels).
Compounded, these two effects can easily explain away $\sim\negmedspace5\%$
differences between two reported systems (if the annotation
noise model is not specified).

Understanding the sensitivity of the model to annotator noise helps
decide how much effort should be put into training the annotators
to be precise versus fast.

\subsubsection{\label{sec:clicks-encoding}Clicks encoding}

Multiple choices are available for encoding corrective clicks as input to $M_{b+c}$. Previous works considered using a distance
transform from the clicks \cite{Xu2016Eccv,Xu2017Bmvc,Liew2017Iccv,Hu2018Nn,Wang2018PamiDeepigeos,Li2018Cvpr},
a Gaussian centred on each click \cite{Le2018Eccv,Maninis2018Cvpr,Mahadevan2018Bmvc},
or a simple binary disk \cite{Bearman2016Eccv} (see supplementary
material for examples). Compared to a binary disk, the distance transform
makes it easier for the convnet to reason about relative
distances to the clicks. The Gaussian might make it easier for
the convnet to localise the exact click centre and to handle cases
where two nearby clicks overlap.

\paragraph{Result}

\begin{figure}
\hfill{}\includegraphics[width=1.05\columnwidth]{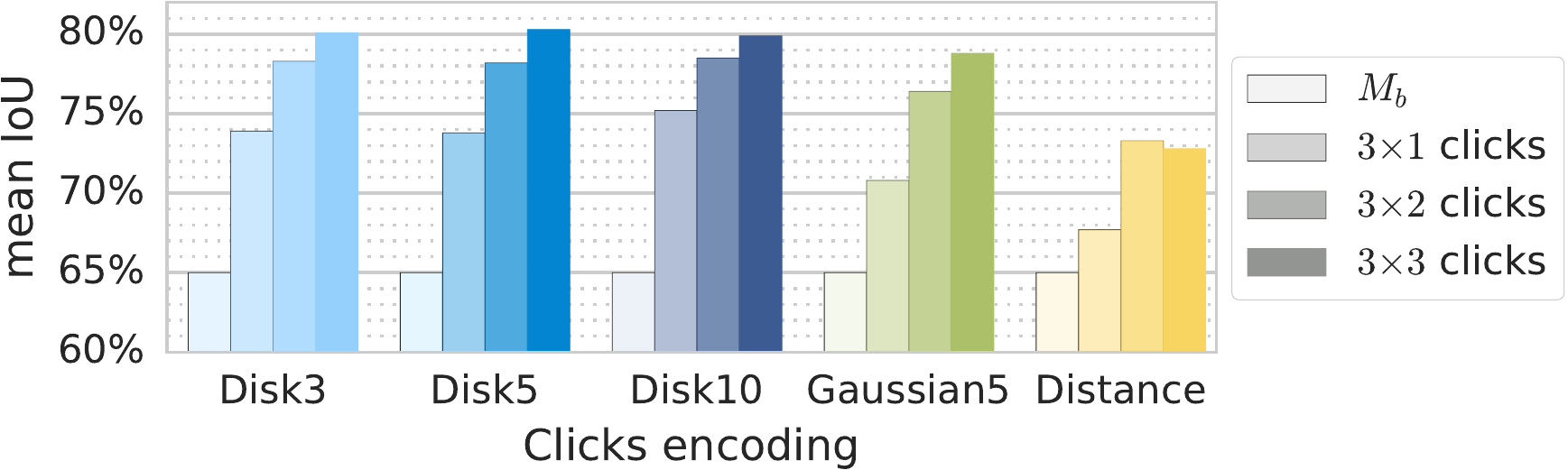}\hfill{}
\vspace{-1.5em}
\caption{\label{fig:clicks-encoding-result}Effect of clicks encoding on the
resulting masks. $M_{b}$ indicates the masks obtained with zero clicks
(bounding box only). Simple binary disks behave better than the alternatives.}
\end{figure}

The results from figure \ref{fig:clicks-encoding-result} indicate
that using a Gaussian or distance transform, surprisingly, underperforms compared
to using a simple binary disk to encode input clicks. The disk
diameter seems make little difference.
In our setup, simplest is best.

We also tried to add the mask generated from the previous round as
an additional input channel, however this did not improve results.

\begin{figure}
\vspace{-1.0em}
\hfill{}\includegraphics[width=1\columnwidth,height=0.65\columnwidth]{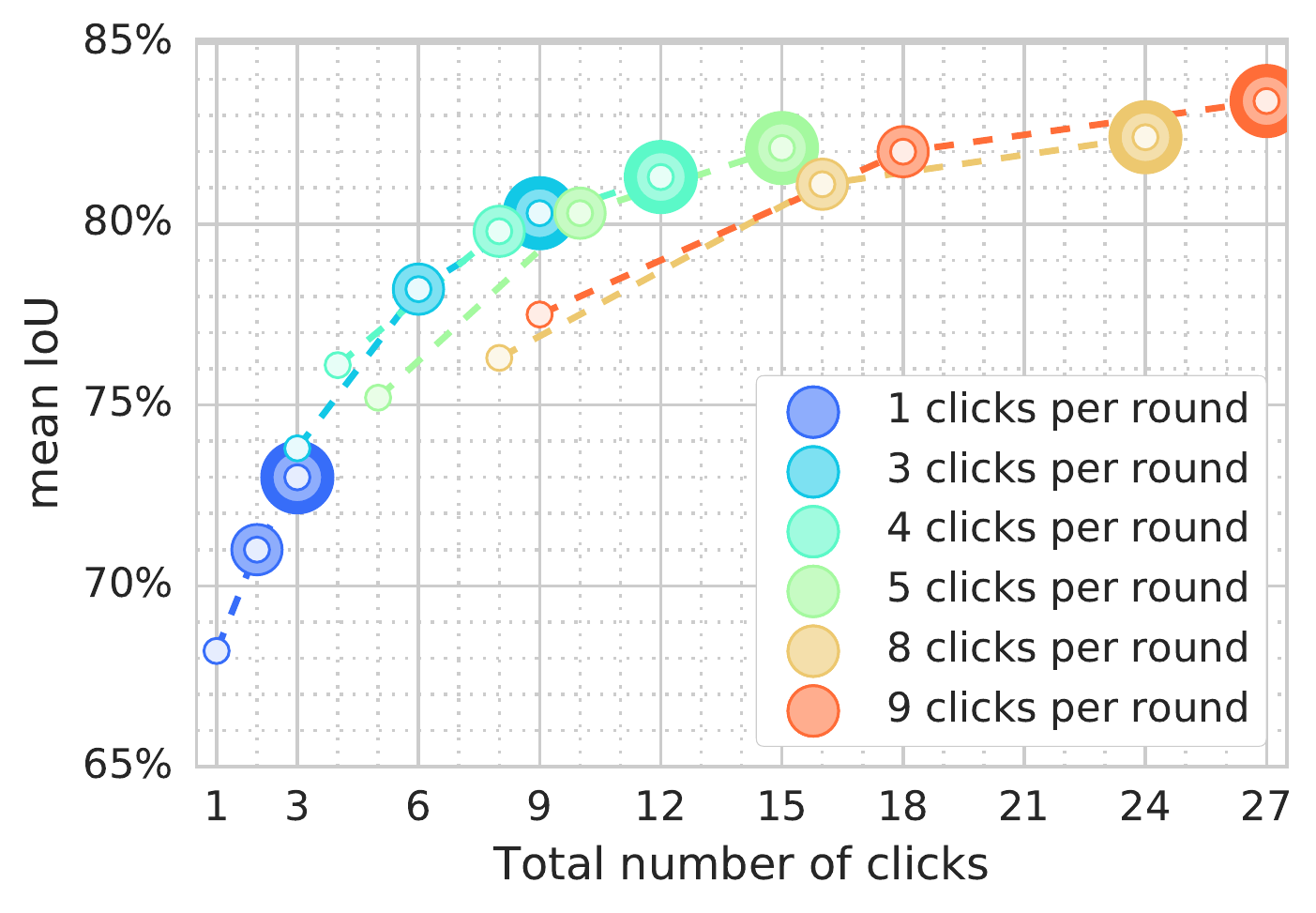}\hfill{}
\vspace{-1.5em}
\caption{\label{fig:number-of-clicks-and-rounds-result}
Effect of varying the number of clicks per round. Each curve show three simulated rounds
(number of rings = rounds done). If \textasciitilde{}9 clicks are to be collected,
it is better to do $3\times3$, $4\times2$, or $5\times2$; than to
do $1\times8$ or $1\times9$. This highlights the benefit of doing
clicks that respond to the model error.}
\vspace{-0.5em}
\end{figure}

\vspace{-0.5em}
\subsubsection{\label{sec:number-of-clicks-and-rounds}Number of clicks and rounds}

Region clicks provide direct supervision for the clicked pixels with
a foreground or background label. The more clicks the better the resulting
masks (with diminishing returns). However it is unclear how to distribute
those clicks across rounds.
As masks are updated inbetween rounds, it might be good
to gather as many clicks as possible per round.
However if too many are done before updating the mask presented to the annotator, we
under-use the extrapolation power of $M_{b+c}$. One click typically
affects the whole region around it and can have global corrective
effect on the generated mask (figure \ref{fig:amazing-corrective-clicks}).

We explore here this trade-off by evaluating three annotation rounds with
different number of clicks per round. They all start from the masks
generated by the $M_{b}$ model ($65\%\ \text{mIoU}$).

\paragraph{Result}

Figure \ref{fig:number-of-clicks-and-rounds-result} shows a clear
gain in mIoU when increasing the total number of corrective clicks,
reaching diminishing returns after \textasciitilde{}15 clicks. At that
point the limits of the $M_{b+c}$ model start to cap the progress
of each click. The figure also shows that if one is, for example, aiming for \textasciitilde{}9
clicks total, it is better to do $3\times3$, $4\times2$, or $5\times2$
clicks; than to do 8 or 9 clicks all in one round. This highlights the
benefits of the interaction between the human annotator and the machine
generating the masks; rather than having the annotator blindly clicking
on all errors of the initial mask.

\subsubsection{\label{subsec:class-agnostic-vs-class-specific models}Class-agnostic or
class-specific?}

Up to now $M_{b}$ and $M_{b+c}$ have been trained in a class-agnostic
manner.
When the target class to annotate is known one might that
suspect that using class-specific models might result in higher quality
masks.

We experimented training car- and giraffe-specific models and evaluate
them over car/giraffes or over all categories. We also trained models
from $\text{ADE}_{\text{L}}^{\text{train}}$ (out-of-domain) as well
as $\text{COCO}_{\text{L}}^{\text{train}}$ (in-domain).

\paragraph{Result}

As expected class-specific $M_{b}$ models generate better initial
results. Similarly in-domain models perform slightly better than out-of-domain
models. However, when adding annotations the gap between models closes
rapidly. After three rounds all variants are within $2$ percent points
$\text{mIoU}$. Even training $\text{ADE}$ or $\text{COCO}$ models
\emph{without} car or giraffe seem to have negligible effect after
three rounds. From these results we conclude that there is no need
to have class-specific models. Instead a model trained with a large
number of instances covering diverse categories performs essentially
just as well.

\section{\label{sec:annotation-process}Large-scale annotation campaign}

Beyond the simulations of section \ref{sec:cc-simulations}, to study
the benefits of corrective clicks we also executed a large-scale annotation campaign over the recently released
OpenImages V4 dataset~\cite{Kuznetsova2018OpenImages}.
This dataset provides bounding box annotations for 600 object categories. We selected
300 categories for which we make instance masks, based on
1) whether the class exhibits one coherent appearance over which a policy
could be defined (e.g. \textquotedbl{}hicking equiment\textquotedbl{}
is rather ill-defined),
2) whether a clear annotation policy can be defined (e.g. which pixels belong to a nose?, see section \ref{sec:policies}),
and 3) whether we expect current segmentation models to be able to capture the shape adequately (e.g. jellyfish contains
thin structures that are hard for state-of-the-art models).
In total, we annotate 2.5M instances over the 300 OpenImages categories considered.
65 of these overlap with the existing 80 COCO categories. We also re-annotate COCO images for these 65 classes, which
allow us to compare with COCO's manual polygon drawing.

In addition to the annotations generated via corrective clicks, we also
make a smaller set of extremely accurate masks fully manually with a free-painting tool (\textasciitilde{}100 instances per class, for a total of 60k masks). We use these as reference for quality evaluation.

Both corrective clicks and the free-painting annotations are made by
a pool of 100 human annotators. These are dedicated full-time annotators, that
have a bias towards quality rather than speed.

Section \ref{sec:corrective-clicks} describes the exact $M_{b}$
and $M_{b+c}$ models setup for collecting corrective clicks.
Section \ref{subsec:Ranking-model} describes the ranking model $M_{r}$.
Section \ref{sec:policies} discusses the annotation policies used,
and section \ref{sec:manual-annotations} describes the free-painting
annotations. We analyse the collected data in section \ref{sec:annotation-analysis}.

\begin{figure}
\hfill{}\includegraphics[width=1\columnwidth,height=0.5\columnwidth]{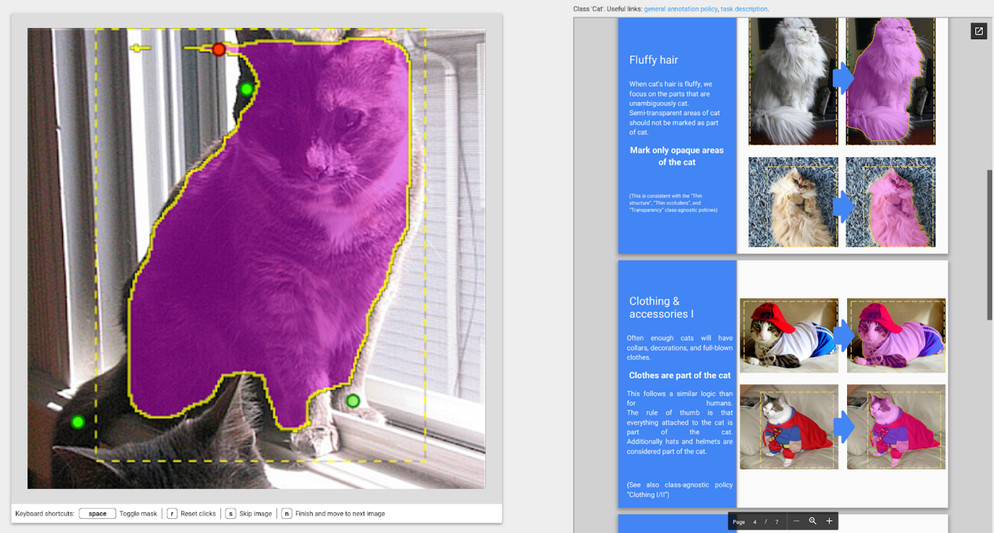}\hfill{}
\vspace{-1.25em}
\caption{\label{fig:cc-ui}The annotation web interface.
Left side shows the object to annotate, the original bounding box (yellow),
the current mask (magenta), and the corrective clicks (green and red dots).
The right side shows the class-specific policy for the class being annotated.}
\end{figure}

\begin{figure}
\vspace{-0.5em}
\hfill{}\includegraphics[width=1\columnwidth,height=0.27\columnwidth]{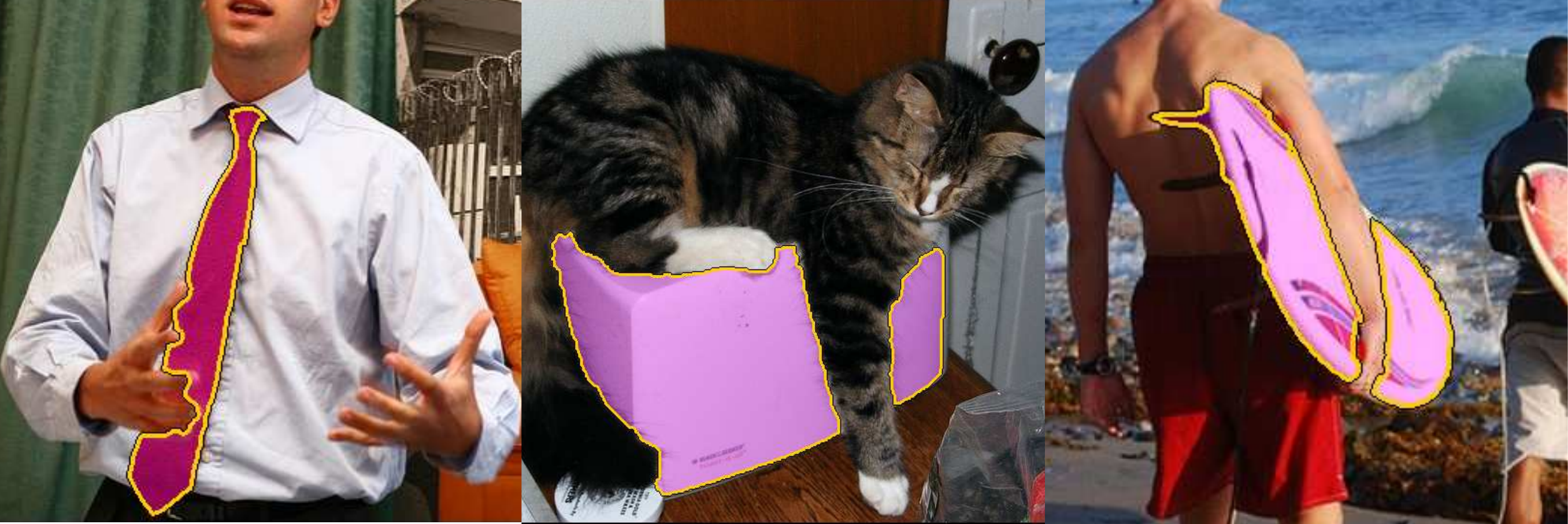}
\vspace{0.2em}
\hfill{}\includegraphics[width=1\columnwidth,height=0.27\columnwidth]{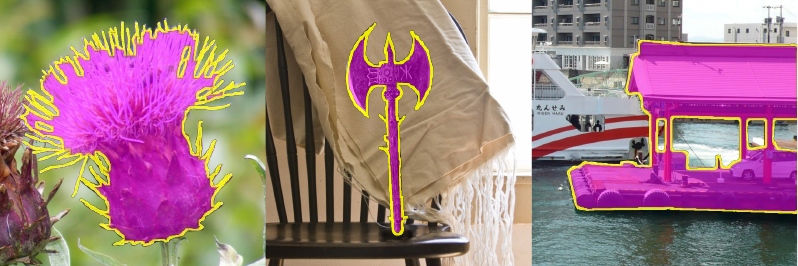}
\hfill{}
\vspace{-1.25em}
\caption{\label{fig:mask-painting-examples}Example of corrective click masks results (top),
and free-painting manual annotations (bottom,  used as ground-truth reference for evaluations).
See supp. material for more examples.}
\end{figure}

\subsection{\label{sec:corrective-clicks}Corrective clicks setup}

For each considered class, we annotate instances from the OpenImages V4
training set with size $\ge\negmedspace80\negmedspace\times\negmedspace40$ (or $\ge\negmedspace40\negmedspace\times\negmedspace80$) pixels.
Based on the simulation results from section \ref{sec:cc-simulations}
we opt to do three rounds of annotations with up to $4$ clicks in each
round (i.e. $4\times3$ setup). Each round is separated by multiple
days. We use section's \ref{subsec:Experiments-blueprint} blueprint
with region clicks, minimal click region
of $10^{2}\ \text{pixels}$%
, and input clicks encoded in two channels (foreground/background clicks) using binary disks of radius $5\ \text{pixels}$.\\
To improve the model quality we:
(1) increased the resolution of the crops fed into $M_{b}$ and $M_{b+c}$
to $513\negmedspace\times\negmedspace513$ pixels (object inside $385\negmedspace\times\negmedspace385$ pixels box);
(2) adjusted the clicks noise to mimic the observed annotators behaviour (i.e. near-uniform clicks inside the error regions);
(3) added a boundary refinement stage based on \cite{Barron2016Eccv}.

\paragraph{Qualification test}

Before starting the main corrective clicks task, each annotator must
first pass an automated qualification test to verify the annotator understands the task,
how to use the web interface (figure \ref{fig:cc-ui}), and that he/she is able to perform
the task in adequate time.
In our experience, using only detailed documentation without an
automated qualification test leads to a slow ramp-up towards producing quality
annotations.

\subsection{\label{subsec:Ranking-model}Ranking model $M_{r}$}

As discussed in section \ref{sec:framework}, we would like to automatically rank
the generated masks from highest quality to lowest quality via a model $M_{r}$.
We train our $M_{r}$ model to regress to the IoU of the ground-truth masks.
We use a random decision forest with five input features:
f1) the number of clicks in the last round;
f2) the round number;
f3) the $\Delta\text{IoU}$ between the mask from previous round and the one from the current round;
f4) the maximum distance between any of the clicks made in the current round and the mask from previous round;
and f5) the average distance between the clicks and said mask.
The regressor is trained using collected annotations over 1\% of $\text{COCO}_{\text{L}}$
instances. We observe that $M_{r}$ training is robust to
the volume of training data and the hyper-parameters of the decision
forest. Once the ranker $M_{r}$ is trained, we apply it over the full
set of COCO and OpenImages corrective click annotations. Examples
of ranked masks can be seen in the supp. material.

Out of the five features used, $\Delta\text{IoU}$ and average click
distance provide the strongest predictive power. $\Delta\text{IoU}$
encodes how much change happened in the last round. Small changes
indicate the mask is already close to its best quality. The average
click distance to the mask is a loose indication of the size of the regions being corrected,
since annotator are instructed to click near the centre of the error regions. The smaller the error region, the higher the mask quality.

Note that our masks ranker relies solely on indirect signals from
the annotation process itself. Since $M_{r}$ does not use class labels
(to avoid capturing class biases) nor image features, it generalises
well across categories and datasets.

\subsection{\label{sec:policies}Annotation policies}

An important consideration when dealing with human annotators is that
the task must be well specified, otherwise miscommunication could
lead to inconsistent outcomes. Compared to drawing bounding boxes,
labelling segmentations has many more ambiguities. Defining annotation
policies is difficult because it is subject to the mismatch between
our self-perceived simplicity of categorisation and the complexity of
the real world\footnote{Simple questions like \textquotedbl{}what are clothes?\textquotedbl{}
are subject of debate for the USA Supreme Court \cite{TheAtlantic2018WhatAreClothes}.}.

We created three types of documents:
1) a manual explaining the task at hand;
2) a class-agnostic policy, which discusses how
to handle transparency, occlusion, fragmented objects, etc.;
3) class-specific policies, which answer questions such as \textquotedbl{}are belts
part of pants?\textquotedbl{}, \textquotedbl{}is the collar part of
a cat?\textquotedbl{}, \textquotedbl{}are ice-cubes in a drink part
of it?\textquotedbl{}.

These policies purely specify what the desired segmentation of an
object should include and exclude. They are defined independently
of the annotation technique. The class-specific policies are defined
aiming to:
a) be feasible for the annotators,
b) result in masks useful to train models,
c) be coherent (so annotators can leverage knowledge built across classes).
The automated qualification tests validates that documents 1\&2 have been understood.
The class-specific annotation policy is shown directly in the annotation
interface for fast consulting (figure \ref{fig:cc-ui}).

In practice defining annotation policies can become a bottleneck for
deployment. They are not trivial to define coherently, it
takes time to find and annotate illustrative examples, and to find
clear concise wording. We thus re-use policies across groups
of similar looking classes (e.g. cat policy for dogs and bears). In
total we created $150+$ slides of policies, defining 42 class-specific
policies covering 200 classes. The class-agnostic policy was
considered sufficient for 100 classes such as frisbee, volleyball,
etc.%

We ran a small scale experiment with 30 novice annotators. After
validating that the task is well understood, we presented common but
non-trivial cases (e.g. bottle with handle, box truck with visible
load). Without providing detailed policies, we observed a near 50/50
split between decision such as \textquotedbl{}is the handle part of
the bottle?\textquotedbl{}, \textquotedbl{}is the load part of the
truck?\textquotedbl{} (see supplementary material). This anecdotal
evidence supports the need for well-defined policies.

\subsection{\label{sec:manual-annotations}Free-painting masks}

We also created a smaller set  of purely
manual annotations over COCO and OpenImages (\textasciitilde{}100 per class). We use these in order
to do evaluations beyond the COCO polygons (which have known inaccuracies).
These annotations are made using the same policies as for corrective clicks.

We provide the annotators with a free-painting tool,
that provides a resizeable circular brush, a \textquotedbl{}straight
lines\textquotedbl{} mode, a \textquotedbl{}fill closed region\textquotedbl{}
mode, erase, and unlimited undo-redo. The free-drawing brush allows
to conveniently delineate curved objects parts (e.g. on animals),
while the \textquotedbl{}straight lines\textquotedbl{} mode allows
to conveniently delineate straight parts (e.g. on man-made objects).
The output of the annotation is not a polygon, but rather a binary
bitmap image.

Like before, we design an automated qualification test, and the annotation
interface shows the per-class policy. We request the annotators to
dedicate about 180 seconds (s) per instance and aim towards near-perfect
quality masks. Examples of the produced masks can be seen in figure
\ref{fig:mask-painting-examples}.

\section{\label{sec:annotation-analysis}Analysis of human annotations}

This section investigates the results of the large scale annotation
campaign described in section \ref{sec:annotation-process}.

\subsection{\label{sec:manual-annotations-analysis}Free-painting annotations}

In total we made free-painting manual annotations for 45k instances of 300
classes on OpenImage,
and 15k instances of 65 classes on COCO (figure \ref{fig:mask-painting-examples}).

\paragraph{Time}

For the large instances considered ($\ge80\times40$ pixels) COCO polygons
have an average of $33.4$ vertices per instance (compared to $24.5$
overall).
Assuming linear time in the number of vertices, we estimate the time taken to
originally annotate COCO with polygons at 108s per large instance (based on the
speed reported in \cite{Lin2014EccvCoco}).
We requested our annotators to spend at least 180s per instance with free-paining. In practice they
took on average 136s per instance.

\paragraph{Quality}

For quality control we double-annotated 5k instances. The average agreement between two
annotations of the same instance is very high at $90\%\ \text{mIoU}$.
This is well above the $\sim\negmedspace80\%\ \text{mIoU}$ human agreement previously
reported for COCO polygons~\cite{Kirillov2018CocoPanopticSlides}.
Moreover, compared to polygons, our annotation tool allows to better annotate
curved objects, our annotators focused on quality rather than speed,
and the resulting masks appear extremely accurate (e.g. pointy tips of the axe, fine details of the flower's boundaries, and thin connecting bars in the boat in Fig.~\ref{fig:mask-painting-examples}).

We conclude from all of the above that our manual free-painting annotations are of even higher quality than the original COCO annotations, and are thus particularly suited for evaluating results of segmentation algorithms.

\subsection{\label{sec:do-annotators-behave}Corrective clicks: Annotators behaviour}
\vspace{-0.25em}

Our $100$
annotators generated $20\text{M}+$ clicks, spread
over $5\text{M}+$ mask corrections.
Let us inspect  these.

\subsubsection{Annotated clicks}
\vspace{-0.25em}

\paragraph{Clicks per round}

For each instance visited in a round the annotators are allowed to
provide 0 to 4 corrective clicks, as well as to click a \textquotedbl{}skip\textquotedbl{} button
to indicate that a mask should not be created
for that instance (according to policy, e.g. because the image is blurry, the object shape is
not visible, the box covers multiple objects, etc.). Zero clicks indicate that
the mask is already good enough and no further corrections are needed.
Neither skips nor 0-clicks masks are sent to the next round. Overall
we observe $2.7\%$ of skips, $2.1\%$ of 0-clicks, and $4.8\%$,
$8.4\%$, $12.3\%$, $70.0\%$ of 1, 2, 3, 4-clicks respectively.
By observing the area distribution of the regions clicked and the masks
IoU (see supp. material) we conclude that annotators under-use the 0-clicks
option and correct minuscule missing details instead. We also observe
they become stricter as rounds progress. %
Besides, we attribute the high percentage of 4-clicks to the annotators' bias towards quality,
and to the fact that it is easier to click on anything wrong (albeit small)
than to judge whether the mask is good enough (see discussion in section~\ref{subsec:Annotation-time}).
Across the three rounds each instance accumulates 3.5, 7.1, and 10.7
clicks on average.

\paragraph{Clicks order}

We expect annotators in each round to first click on large error regions, and
then on smaller ones. For each annotated instance we computed the area of the error region
corresponding to each click. We observe that $60\%$ of the clicks are in approximately large-to-small order (and $30\%$ are exactly so). Additionally, we observe that the average area of the 1st, 2nd,
3rd, and 4th clicked error region are in strictly decreasing order (see supplementary material).
Overall, annotators click first the largest error region and then proceed to finer details.

\paragraph{Clicks distribution}

Annotators are instructed to do one click per error region, unless
it is rather big, in which case the clicks should be spread
across the region. We measure the area of the error region as a function of number
of received clicks. We observe that indeed only the smallest regions
are left without clicks, and that the number of clicks grows almost
linearly with the area of the error region (at about \textasciitilde{}$22^{2}\ \text{pixels}$
per click). About
$80\%$ of the clicked regions received 1 click, $15\%$ 2 clicks,
$4\%$ 3 clicks, $1\%$ 4 clicks.
Overall, annotators indeed only do multiple clicks if the region to correct is rather large.

\vspace{-1em}
\subsubsection{\label{subsec:Annotation-time}Annotation time}

\paragraph{Time per instance}

The annotation interface shows both the instance to annotate as
well as the class-specific policy (figure \ref{fig:cc-ui}). The
motion of the mouse is continuously logged. If we measure the time
spent on the annotation interface we obtain an average of $11.4$s
per instance (averaged over all classes and rounds). We observe a
significant variance across classes: the fastest 0.1 quantile of the classes
take $<8.7$s per round on average, while the slowest 0.9 quantile
averages $>12.7$s per round.

\paragraph{Time vs number of clicks}

\begin{figure}
\begin{minipage}[c][1\totalheight][t]{0.45\columnwidth}%

\vspace{-1em}
\hfill{}\includegraphics[width=1\columnwidth]{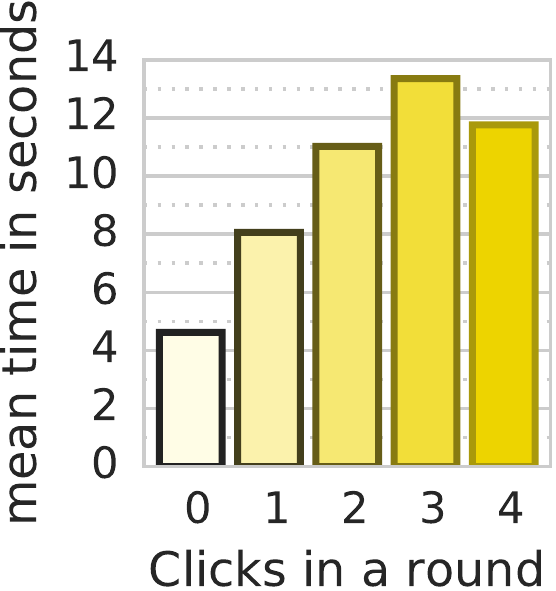}\hfill{}

\caption{\label{fig:time-per-answer-type}Mean time in seconds per answer type.
See section \ref{subsec:Annotation-time} for discussion.}
\end{minipage}\hfill{}%
\begin{minipage}[c][1\totalheight][t]{0.51\columnwidth}%

\vspace{-1em}
\hfill{}\includegraphics[width=1\columnwidth]{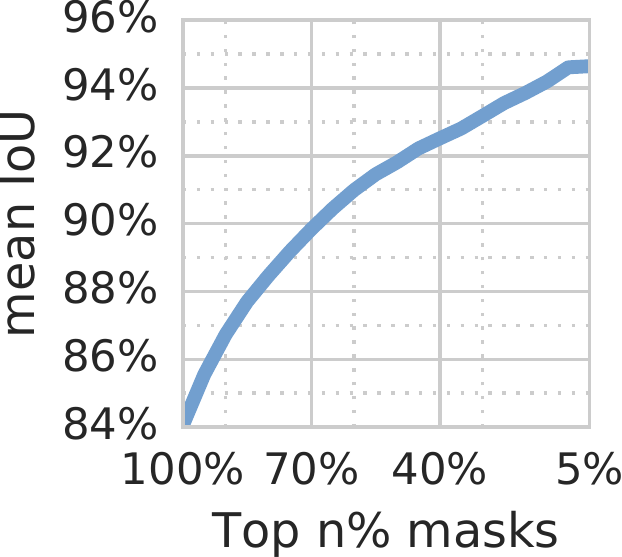}\hfill{}

\caption{\label{fig:miou-vs-fraction-top-ranked}mIoU over $\text{COCO}_{\text{L}}$
for fraction of top ranked samples. $M_{r}$ allows to select high
quality subsets of the data.}
\end{minipage}
\vspace{-0.5em}
\end{figure}

Figure \ref{fig:time-per-answer-type} shows the average time as a function of the number of clicks in a round.
Deciding if the mask is good enough (0-clicks) takes about 4s, after which the average time continuously grows until 3 clicks. Curiously, doing 4 clicks is faster than doing 3.
We hypothesize this is because masks with large obvious errors need all 4 clicks, and
these can be done fast, whereas a 3 click answer requires extra time to
decide to withhold the 4th click.
Furthermore, the delay between the moment the instance is shown and the first action
is taken suggests that annotation time can be split between \textquotedbl{}scene
parsing\textquotedbl{} and \textquotedbl{}click actions\textquotedbl{}.
After the first click is done (8s), it takes about 3s per additional click.

\paragraph{Time vs area}

For first clicks done under 10s, we see a direct relation between speed and
the area of the clicked error region.
Faster first clicks are done on larger areas ($41^2$ pixels for a \textasciitilde{}3s first click), and slower first clicks are done on smaller areas ($33^2$ for a \textasciitilde{}8s first click).
For first clicks done above 10s there is no clear relation. These are cases where annotators
wonder what should be clicked.

\paragraph{Time per round}

The average time increases over rounds, with $10.8$, $11.7$,
and $11.9$ seconds for rounds 1, 2, 3, respectively. This is
consistent with the mask errors becoming smaller and thus taking more
time to find.

\vspace{-0.25em}
\subsection{\label{sec:time-vs-masks-quality}Corrective clicks: Time versus
quality}
\vspace{-0.25em}

\begin{figure}
\vspace{-0.7em}
\hfill{}\includegraphics[width=0.98\columnwidth]{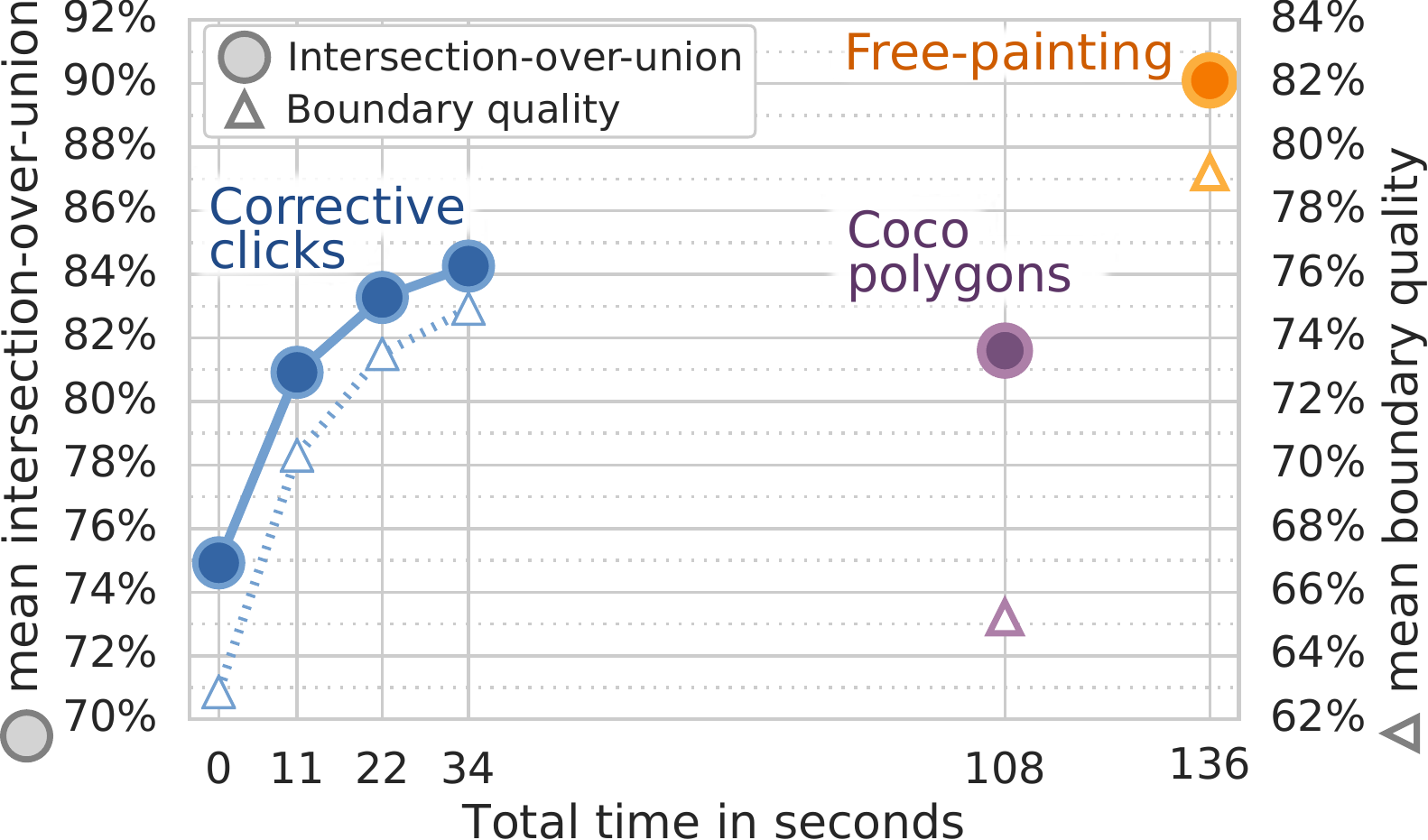}\hfill{}
\vspace{0em}
\caption{\label{fig:coco-uber-miou-vs-time}
Masks quality versus time for different annotation schemes (section \ref{sec:time-vs-masks-quality}).
Our corrective click masks reach better quality than COCO polygons, while being $3\times$ faster to make.}
\vspace{-0.5em}
\end{figure}

Each annotation method presents a different trade-off between speed
and quality. Fig. \ref{fig:coco-uber-miou-vs-time} summarizes
this for our corrective clicks, COCO polygon annotations, and
our free-painted masks. In all cases mIoU is measured w.r.t. the free-painted annotations on the COCO dataset (\S \ref{sec:manual-annotations} and \ref{sec:manual-annotations-analysis}).

After three rounds of corrective clicks, we obtain masks with $84\%\ \text{mIoU}$ in an average human annotation time of $34\text{s}$ per instance. In comparison, COCO polygons reach $82\%\ \text{mIoU}$, while taking an estimated $108\text{s}$ per instance (section \ref{sec:manual-annotations-analysis}).
For reference, our free-painted masks take $138\text{s}$ and have
a self-agreement of $90\%\ \text{mIoU}$.
We observe a similar trend when comparing boundary quality (F-measure at 5 pixels \cite{Martin2004Pami, Pont2016supervised}): 75\% for our corrective click masks versus 65\% for COCO polygons, and 79\% for the reference free-painted masks.
\emph{We thus conclude that our corrective click masks are of higher
quality than the COCO polygon annotations, while being $3\times$ faster to
make}.

Coincidentally, COCO annotations average $33.4$ vertices
per instance on the instances considered, and our annotator do
an average of $10.7$ clicks per instance over all three rounds of interactive segmentation.
Thus the $3\times$ factor also holds in number of clicks.
Some examples of our corrective click masks can be seen in figure \ref{fig:mask-painting-examples}.

\paragraph{Semantic segmentation training}
To further validate the utility of the generated corrective click
masks, we train from them a DeeplabV3 Xception65 model for semantic segmentation
over $\text{COCO}^{train}_{L}$.
For comparison, we also train a second model from the original COCO polygon annotations.
Since we only annotated large objects ($>80\times40$ pixels), we ignore small instances during both training and testing.
We then evaluate over $\text{COCO}_{\text{L}}^{\text{test}}$ and observe
that both models perform comparably ($52\%\ \text{mIoU}$ COCO masks, versus
$53\%$ ours).

\vspace{-0.25em}
\subsection{\label{sec:is-the-ranking-good}Corrective clicks: Masks ranking}
\vspace{-0.25em}

Once all our annotations are produced, the $M_{r}$ model can be used
to rank them. We train this model on $1\%$ of COCO ground-truth (section \ref{subsec:Ranking-model})
and then use it to rank all corrective click masks.

Fig. \ref{fig:miou-vs-fraction-top-ranked} shows mIoU over
$\text{COCO}_{\text{L}}$ when selecting the top N\% ranked masks
(bottom N\% plot in supp. material). The slanted shape indicates
that $M_{r}$ is effective at sorting the masks. A random ranker would
result in a horizontal line. Thanks to $M_{r}$ we can select a higher
quality subset of the data (the top 70\% masks have $90\%\ \text{mIoU}$),
target annotation efforts on the lowest ranking instances (the
bottom 30\% have $70\%\ \text{mIoU}$), or weight training samples
based on their rank. This self-diagnosing capability is a side-effect of
using corrective clicks rather than directly drawing the masks.

\section{Conclusion}

We have shown that interactive segmentation can be a compelling approach for instance segmentation at scale.
We have systematically explored the design space of deep interactive segmentation models. Based on the gained insights, we executed a large-scale annotation campaign, producing 2.5M instance masks on OpenImages.
These masks are of high quality (84\% mIoU, 75\% boundary quality).
Additionally, we proposed a technique for automatically estimating the quality of individual masks.
We plan to publicly release these new annotations hoping they will help further develop the field of instance segmentation.

\FloatBarrier

\bibliographystyle{ieee}
\bibliography{2019_cvpr_corrective_clicks}

\end{document}